\documentclass[10pt,twocolumn,letterpaper]{article}

\usepackage{cvpr}
\usepackage{times}
\usepackage{epsfig}
\usepackage{graphicx}
\usepackage{amsmath}
\usepackage{amssymb}
\usepackage{booktabs}
\usepackage{gensymb}
\usepackage{lib}
\usepackage{wrapfig}

\usepackage[pagebackref=true,breaklinks=true,letterpaper=true,colorlinks,bookmarks=false]{hyperref}

 \cvprfinalcopy 

\def\cvprPaperID{3257} 

\ifcvprfinal\fi
\begin{document}

\title{Learning Independent Object Motion from Unlabelled Stereoscopic Videos}

\author{Zhe Cao\\
UC Berkeley\\
{\tt\small zhecao@berkeley.edu}
\and
Abhishek Kar\\
Fyusion Inc.\\
{\tt\small akar@fyusion.com}
\and
Christian Haene\\
Google\\
{\tt\small chaene@google.com}
\and
Jitendra Malik\\
UC Berkeley\\
{\tt\small malik@eecs.berkeley.edu}
}

\maketitle

\begin{abstract}
We present a system for learning motion of independently moving objects from stereo videos. The only human annotation used in our system are 2D object bounding boxes which introduce the notion of objects to our system. Unlike prior learning based work which has focused on predicting dense pixel-wise optical flow field and/or a depth map for each image, we propose to predict object instance specific 3D scene flow maps and instance masks from which we are able to derive the motion direction and speed for each object instance. Our network takes the 3D geometry of the problem into account which allows it to correlate the input images. We present experiments evaluating the accuracy of our 3D flow vectors, as well as depth maps and projected 2D optical flow where our jointly learned system outperforms earlier approaches trained for each task independently.
\end{abstract}

\section{Introduction}

Considering the crowded road scene in \figref{teaser}, what information do we as humans use to navigate effectively in this environment? We need to have an understanding of the structure of the environment, i.e. how far other elements in the scene (cars, bikes, people, trees) are from us. Moreover, we also require knowledge of the speed and direction in which other agents in the environment are moving relative to us. Such a representation, in conjunction with our ego-motion, enables us to produce a hypothesis of the environment state in the near future and ultimately allows us to plan our next actions.

In order to gather this information, humans use stereo-motion, i.e. a stream of images captured with our two eyes as we move through the environment. In this work, we develop a computational system that aims to produce such a factored scene representation of 3D structure and motion from a binocular video stream. Specifically, we propose to predict the 3D object motion of each moving object (represented by 3D scene flow) in addition to a detailed depth map of the scene from a stereo image sequence. This task and its variants have been tackled in supervised settings which require labels such as dense depth maps and motion annotations that are prohibitively expensive to collect or alternatively obtained from synthetic datasets~\cite{eigen2014depth,fischer2015flownet, ilg2017flownet, kendall2017end,mayer2016large}. We present a system that learns to predict these quantities using only unlabelled stereo videos, thus making it applicable at scale. In addition to producing pixel-wise depth and scene flow maps, our network is aware of the notion of independent objects. This allows us to produce a rich factored 3D representation of the environment where we can measure velocities of independent objects in addition to their 3D positions and extents in the scene. The only labels used by our system are those introduced by off-the-shelf object detectors which are very cheap to acquire at scale.

\begin{figure}[t]
    \centering
    \includegraphics[width=0.48\textwidth]{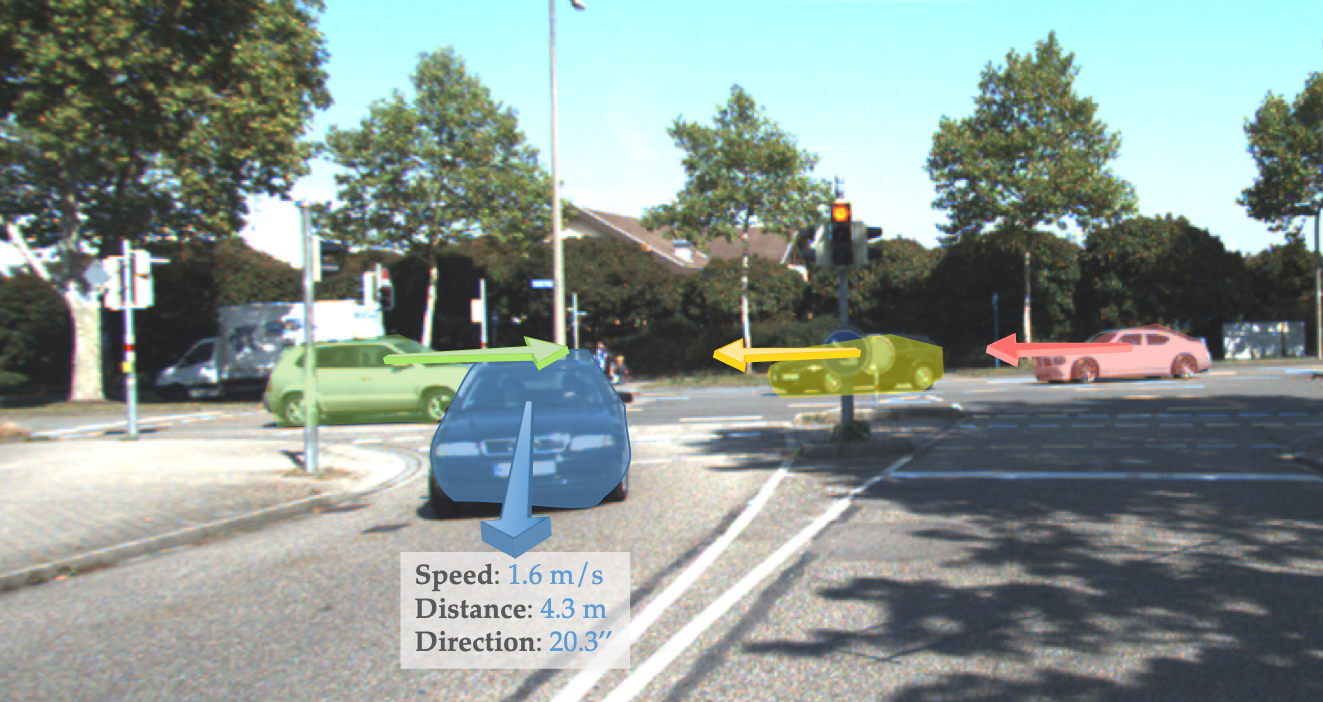} \\
    \caption{Object motion predicted by our system. Trained with raw stereo motion sequences in a self-supervised manner, our model learns to predict object motion together with the scene depth using sequence of stereo images and object proposals as input. The speed and moving direction of each moving object is derived from our scene flow prediction.}
    \figlabel{teaser}
    \vspace{-5pt}
\end{figure}

 
 Prior work in this domain has focused on certain sub-problems such as learning depth or optical flow prediction without explicit labels~\cite{yin2018geonet,godard2017unsupervised,garg2016unsupervised}. In \secref{experiments}, we demonstrate that by jointly learning the full problem of depth and scene flow prediction, we outperform these methods for each of these sub-problems as well. The key contributions of our work are as follows: (1) formulating a learning objective  which works with the limited amount of supervision that can be gathered in a real world scenario (object bounding box annotations), (2) factoring the scene representation into independently moving objects for predicting dense depth and 3D scene flow and (3) designing a network architecture that encodes the underlying 3D structure of the problem by operating on plane sweep volumes.


The following sections are organized as follows. \secref{related_work} discusses prior work on inferring scene structure and motion. \secref{method} presents our technical approach for inferring scene flow from stereo motion - loss functions, object centric prediction and priors. In \secref{network}, we describe our network architecture designed for geometric matching and 3D reasoning in plane sweep volumes. \secref{experiments} details our experiments on the KITTI dataset \cite{Menze2015CVPR} with extensive evaluation of our depth and scene flow prediction.

\begin{figure*}[t]
    \centering
    \includegraphics[width=0.85\linewidth]{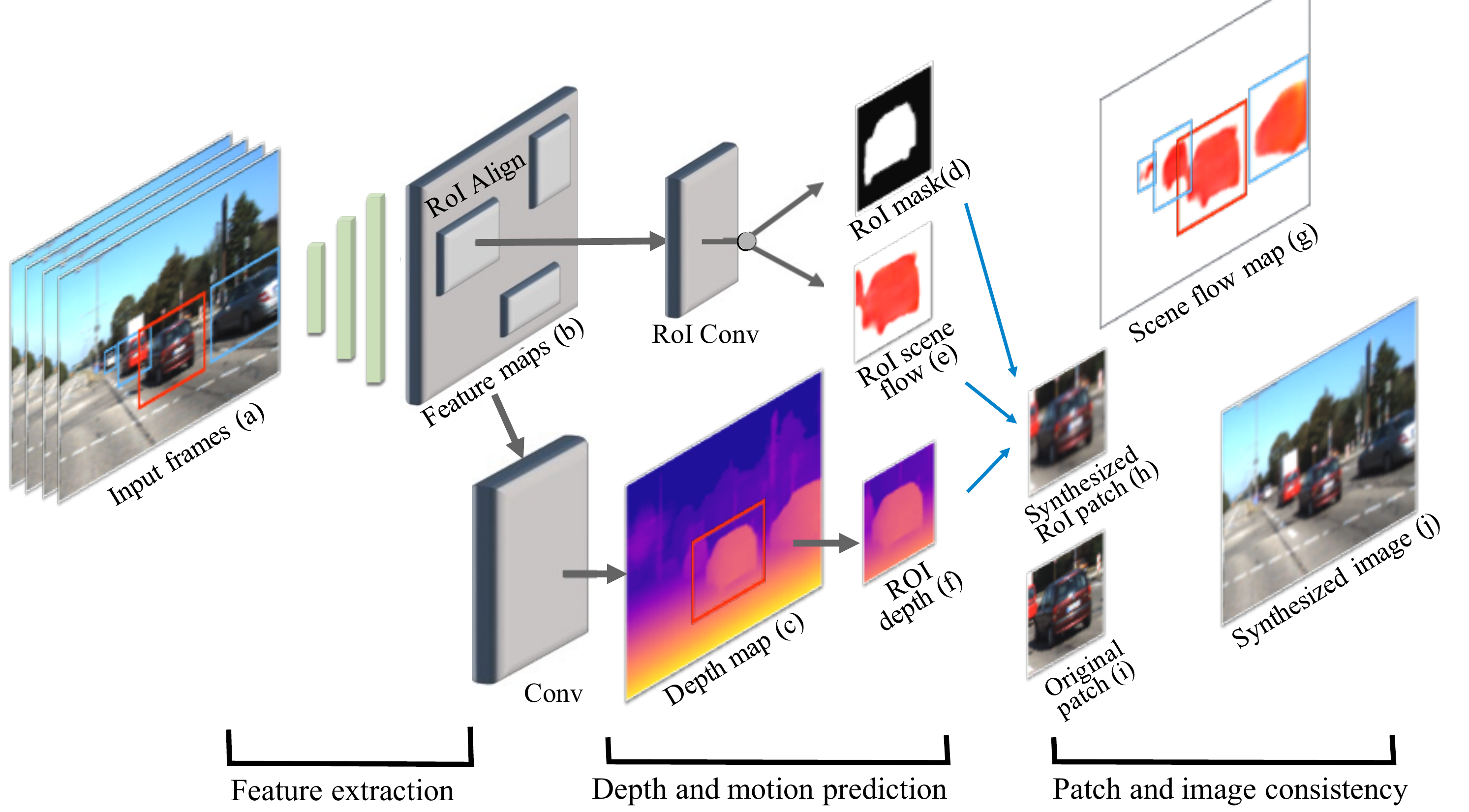} \\
    \caption{Our pipeline for learning depth and object motion from a stereo motion sequence. Using a stereo motion sequence as input, our system predicts a depth map (c) for the reference image, instance mask (d) and 3D scene flow (e) for each independent moving object in a single forward pass. Using the instance mask and scene flow prediction, we can compose a full scene flow map (g). For each RoI region, we synthesize a RoI patch (h) based on the ROI camera intrinsics, ROI depth (f), 3D scene flow (d) and instance mask (e) as explained in \secref{selfsup}. We use the synthesized patch (h) and original patch (i) from the input image to enforce consistency losses to supervise the RoI prediction. We use stereo reprojection to supervise the depth prediction. Finally, we use the full map scene flow and depth to synthesize a image (j) for computing the consistency loss with the input image.}
    \figlabel{pipeline}
\end{figure*}

\section{Related work}
\seclabel{related_work}

In our work we recover scene geometry and object motion jointly while traditionally these problems have been solved independently.
The geometry of a scene is reconstructed by first recovering the relative camera pose between two or more images taken from different viewpoints using Structure-from-Motion (SfM) techniques \cite{longuet1981computer,hartley2003multiple}. Using a subsequent dense matching and triangulation we can eventually recover a dense 3D model of the scene \cite{pollefeys2004visual}. The underlying assumption within the aforementioned methods is that the scene is static, i.e.\ does not contain any moving objects. The case for independently moving objects has also been studied in a purely geometric setting \cite{costeira1998multibody}. The key difficulties are degenerate configurations and outliers in point correspondences \cite{ozden2004reconstructing}. Therefore additional priors are used - a common example is objects moving on a ground plane \cite{yuan20063d}. Similarly, estimating the shape of non-rigid objects is ambiguous and hence using additional constraints such as maximizing the rigidity of the shape \cite{ullman1984maximizing} or representing the non rigid shape as linear combination of base shapes \cite{bregler2000recovering} have been proposed. When reconstructing videos captured in unconstrained environments additional difficulties such as incomplete feature tracks and bleeding into the background have to be handled \cite{fragkiadaki2014grouping}. Our proposed approach is trained on real world data which makes it robust to appearance variations and suitable priors are directly learned from data.

Vedula~\etal.~\cite{vedula1999three} introduced the problem of 3D scene flow estimation, where for each point a 3D motion vector between time $t$ and $t+1$ is computed. Different variants are considered depending on the amount of 3D structure that is given as input. A common variant is to consider a stream of binocular image pairs of a moving camera as input \cite{huguet2007variational,wedel2008efficient,vogel2013piecewise,Menze2015CVPR,taniai2017fast}, and give a depth and 3D scene flow as output. This is often referred to as the stereo scene flow estimation problem. Similarly RGBD scene flow considers a stream of RGBD (color and depth) images as input \cite{jaimez2015primal}.

Recently learning-based approaches, especially convolutional neural networks have been applied for single view depth prediction \cite{ladicky2014pulling,eigen2014depth}, optical flow \cite{fischer2015flownet}, stereo matching and scene flow \cite{mayer2016large}. These learning systems are trained using ground truth geometry and/or flow data. In practice such data is only available for synthetic data in a large scale. A natural way to complement the limited amount of ground truth data is using weaker supervision. For the aforementioned problems, loss functions which are purely based on images and rely on photometric consistency as learning objective have been proposed \cite{garg2016unsupervised,zhou2017unsupervised,godard2017unsupervised,tulsiani2017multi,vijayanarasimhan2017sfm}. They essentially utilize a classical non-learned system \cite{furukawa2010accurate} within the loss function. A few recent works \cite{yin2018geonet, zou2018df, yang2018every, luo2018every, ranjan2018adversarial} use such a self-supervised approach to predict optical flow and depth. To our knowledge our work is the first network that learns to directly predict object specific 3D scene flow without relying on pixel-wise flow or depth annotations.

Another key difference of our work from prior works that predict depth and optical flow is that they predict depth based on a single image. This limits their performance as demonstrated in our results. Geometric reasoning can be included into the network architecture as demonstrated in \cite{kendall2017end,kar2017learning,ji2017surfacenet,yao2018mvsnet}.  We extend these ideas to full 3D scene flow estimation while also operating at the level of object instances allowing us to produce rich factored geometry and motion representation of the scene.

\section{Scene Flow from Stereo Motion}
\seclabel{method}
\subsection{Problem Setup}


\figref{pipeline} illustrates the design of our system. The input to our system is a stream of calibrated binocular stereo image pairs $\mathcal{I} = \{I_1^l,I_1^r, \ldots I_{n}^l, I_{n}^r\}$ captured from times $1$ to $n$.  The most common case we are investigating is that of $n=1$, i.e. two consecutive binocular frames at time $t$ and $t+1$. The intrinsic camera calibration $K$ is assumed to be known. The camera poses of the left camera at each time instant are denoted by $\mathcal{T} = \{T_1,\ldots,T_{n}\}$ and are precomputed using visual SLAM \cite{Geiger2011IV}. For any given time instant $t$, we also have a set of $j$ 2d bounding box detections (or RoIs) $\mathcal{B} = \{ B^1, \ldots, B^j \}$ on the left image $I_{t}^l$ predicted by an off the shelf object detector. The task is to compute the following quantities for the reference frame - a dense depth map $D$, a set of dense 3D flow fields $\mathcal{F} = \{ F^1,\ldots,F^j \}$ that describe the motion between $t$ and $t+1$ and a set of instance masks $\mathcal{M} = \{ M^1,\ldots,M^j\}$ containing all the pixels for each moving object. From these instance-level predictions, we can compose the full scene flow map $F$ by assigning a 3D scene flow vector to each image pixel in the full image. 


We design our system as a convolutional neural network (CNN) which learns to predict all quantities jointly and train the network in a self-supervised manner. The supervision comes from the consistency between synthesized images and input images at different time instants and from different camera viewpoints. The basic principle is that given the predictions of the scene flow $F$ and depth $D$ in a frame $I_\mathrm{ref}$, we can use the precomputed ego-motion to warp another image $I$ into the reference view. This process generates a synthesized image which we call $\hat{I}$. We then define our learning objective as the similarity between the captured images $I_{\mathrm{ref}}$ and the synthesized images $\hat{I}$. The above principle is then applied to RoIs independently followed by an assembly procedure for full image scene flow. This allows us to produce a factored representation of the environment into static and dynamic objects with high quality estimates of object extents (instance masks), depth and motion.

\subsection{Disentangling Camera and Object Motion}
The scene motion in a dynamic scene captured by a moving camera can be decomposed into two major elements - the motion of static background resulting from the camera motion and the motion of independently moving objects in the scene. A common way to represent the scene motion is 2D optical flow. However, this representation confounds the camera and object motion. We model the motion of the static background using the 3D structure represented as a depth map and the camera motion. Dynamic objects are modelled with full 3D scene flow. To this end, we utilize 2D object detections in the form of bounding boxes from an off-the-shelf object detection system and reason about the 3D motion of each object independently. We present a method to learn such a representation for each object proposal in a self-supervised manner by leveraging photo-consistency.

\subsection{Supervising Scene Flow by View Synthesis}
\seclabel{selfsup}

The key supervision for the scene flow prediction comes from the photometric consistency of multiple views of the same scene. The process is illustrated in ~\figref{depth_consist}. Our network predicts a depth map $D$ and a scene flow map $F$ for the reference view $I_\mathrm{ref}$. Using a different image $I$ we can use the predictions to warp $I$ into the reference view and generate a synthesized image $\hat{I}$. We then minimize the photometric difference between $I_{\mathrm{ref}}$ and $\hat{I}$ given as
\begin{multline}
\label{eq:pho}
    \mathcal{L}_\mathrm{photo} = \\ \alpha\frac{1-\operatorname{SSIM}(I_\mathrm{ref},\hat{I})}{2}+(1-\alpha)\|I_\mathrm{ref}-\hat{I} \|_1,
\end{multline}
using the robust image similarity measurement from~\cite{godard2017unsupervised}.

We denote the homogeneous coordinates of pixel $p$ as $h(p)$. A pixel $p$ from the reference frame is transformed to a pixel $\hat{p}$ within a frame $I$
\begin{equation}
    \label{eq:proj}
h(\hat{p}) = KT_{\mathrm{rel}}(D(p)K^{-1}h(p) + F(p)),
\end{equation}
with $T_{\mathrm{rel}}$ the relative transformation from reference frame to $I$. This allows us to do a reverse warp using bilinear interpolation, keeping the formulation differentiable. 

\begin{figure}[tb]
    \centering
    \includegraphics[width=1.0\linewidth]{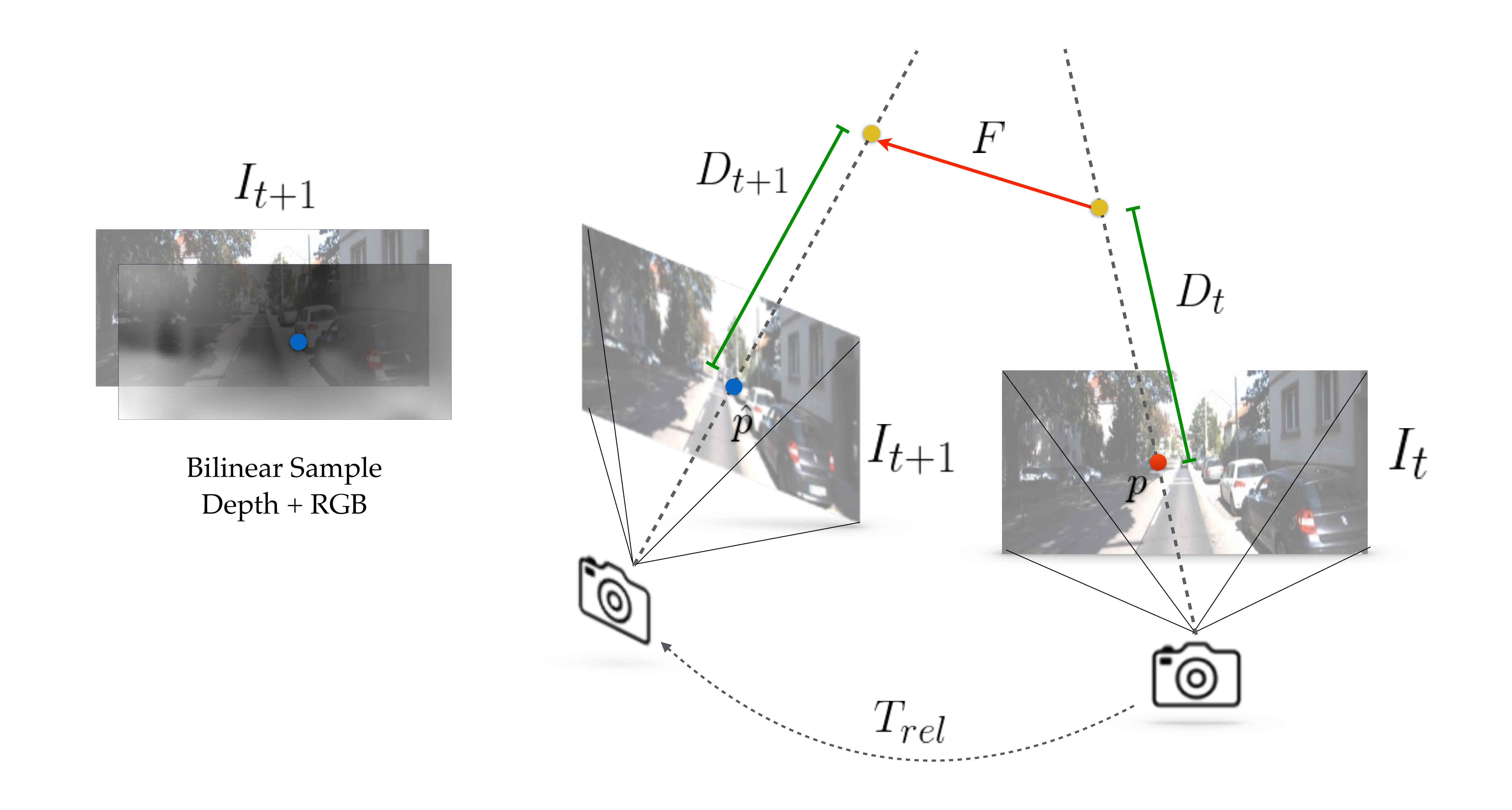} \\
    \caption{Illustration of our image reprojection process. A pixel $p$ from image $I_{t}$ is unprojected using its predicted depth and subsequently transformed to the frame of $I_{t+1}$ using the predicted flow $F$ and the camera transform $T_{rel}$. The photometric consistency loss is derived from the photometric difference between $I_t$ and $\hat{I}_{t+1 \arrow t}$ where $\hat{I}_{t+1 \arrow t}$ is created by warping $I_{t+1}$ into $I_{t}$. The geometric consistency loss is computed by comparing the difference between depth maps warped in the above manner and having them consistent with the z-dimension of the predicted flow $F$. Note that using only photometric consistency would not resolve the ambiguity in the $z$ direction of the flow.}.
    \vspace{-10pt}
    \figlabel{depth_consist}
\end{figure}

Using the photometric consistency alone is insufficient for supervising the 3D flow prediction. The reason is that along a viewing ray multiple photo consistent solutions are possible, as shown in ~\figref{depth_consist}. Therefore we use an additional geometric loss in form of a depth consistency which further constrains the flow. The idea is that the flow in $z$-direction, sometimes also called disparity difference has to agree with the depth maps predicted for the two time instants $t$ and $t+1$. In order to utilize this loss function a depth map for both time instants needs to be predicted and the warping is applied to the depth map.

Analogous to the photometric consistency, the geometric consistency is defined by comparing the depth values of the warped image and reference image. It is then defined as
\begin{equation}
\label{eq:depth_con}
    \mathcal{L}_\mathrm{geo} = \left\|D_\mathrm{ref} - \hat{D} + F_z\right\|_1.
\end{equation}
where $F_z$ is the z-dimension of the predicted sceneflow.
As we have a pair of stereo frames at two time instants, we use this form of warping in two ways - (1) warping between the left and right frames $\{I_{t}^l, I_{t}^r\}$ at a given time instant $t$ to supervise the depth \cite{garg2016unsupervised} and (2) warping between the left or right frames at times $t$ and $t+1$ to supervise both depth and flow. Note that in case of the former, the scene flow vectors are zero as both images are from the same time instant.

\subsection{Object-centric Scene Flow Prediction}

Image based consistency losses are typically applied by warping the whole image and then computing the consistency over the whole image - examples for optical flow prediction can be found in~\cite{yin2018geonet,zou2018df}. For 3D scene flow this is not an ideal choice due to the sparsity of non-zero flow vectors. Compared to the static background, moving objects constitute only a small fraction of the image pixels. This unbalanced moving/static pixel distribution makes naively learning full image flow hard and ends up in zero flow predictions even on moving objects. To make the network focus on predicting the correct flow on moving objects and provide a more balanced supervision, we therefore use object bounding box detections obtained from a state-of-the-art 2D object detection system~\cite{liu2016ssd}. It is important to note that the object detection does not actually tell us if the object is moving or not. This information is learned by our network using our view synthesis based loss functions.

Formally each flow prediction happens in a region of interest (RoI) within the original image, with size and location $B = \big[x, y, w, h\big]$. In our system the per-object flow map is predicted at a fixed size $w_r \times h_r$ using a RCNN based architecture as detailed in \secref{network}. In order to use our view synthesis based loss functions we need to transform the camera parameters which are given for the whole image into RoI specific versions. The change only affects the intrinsic camera parameters and hence we need to compute a new intrinsic matrix $K^j$ for each RoI $j$. The transformation ends up to be a displacement of the principal point and scaling of the focal length.
\begin{figure}[t]
    \centering
    \includegraphics[width=0.35\textwidth]{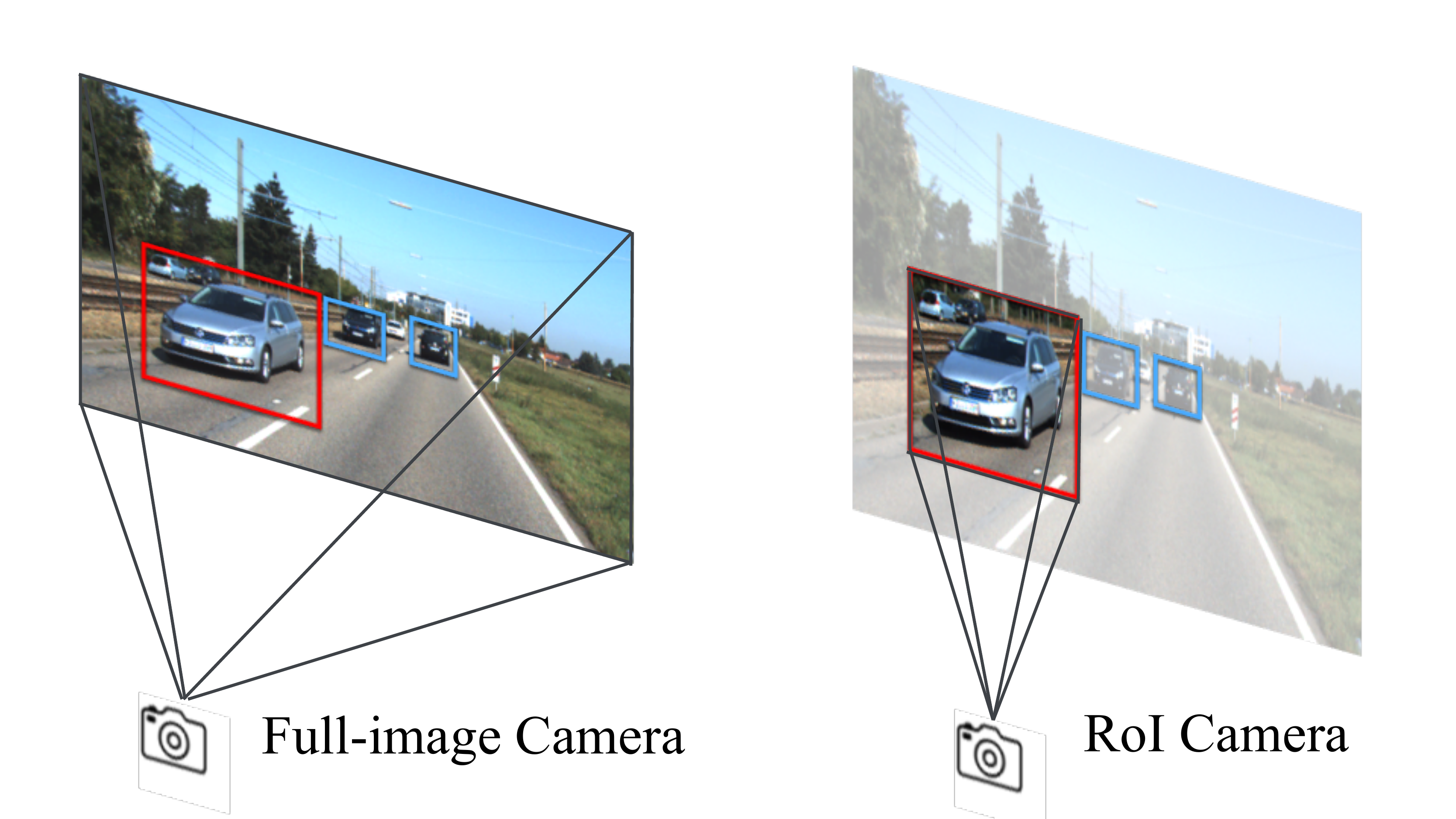} \\
    \caption{Illustration of image rescale and crop process and the change in the camera intrinsics.}
    \label{fig:roialign}
    \vspace{-5pt}
\end{figure}
\begin{align}
K = &
\begin{bmatrix}
f_x & 0 & c_x \\
0 & f_y & c_y\\
0 & 0 & 1\\
\end{bmatrix}, \\
K^j = &
\begin{bmatrix}
{f_x}{w_r}/{w} & 0 & (c_x-x){w_r}/{w} \\
0 & {f_y}{h_r}/{h} & (c_y-y){h_r}/{h}\\
0 & 0 & 1\\
\end{bmatrix}
\end{align}

Note that we do not need bounding box associations between different viewpoints or time instants. We only compute detections for frame $I_t^l$ and use a slightly expanded area as our RoI in frames that we warp to our reference frame for computing consistency losses in Eq.~\ref{eq:pho} and \ref{eq:depth_con}.

\subsection{RoI Assembly for Full Frame Scene Flow}
\seclabel{roiassembly}
We assemble a complete scene flow from the object specific maps $F^j$. However, we have overlapping RoIs and certain RoIs may even contain multiple moving objects. Thus predict an object mask $M^j$ for each RoI $j$ in addition to $F^j$. The full 3D scene flow map $F$ is computed as:
\begin{equation}
    F = \textstyle\sum_{j} M^j \odot F^j.
\end{equation}
We then use the full image flow map $F$ with Eq.~\ref{eq:pho} and Eq.~\ref{eq:depth_con} for full image photometric and geometric losses. Note that the assembly procedure is fully differentiable and we are able to train instance masks $\mathcal{M} = \{ M^1,\ldots,M^j\}$ without any explicit mask supervision. We later use these instance masks (with flow) to identify moving objects (cf.\figref{exp_mask}).

\subsection{Full Learning Objective}

We first state our full image synthesis based loss and then explain further priors we impose in our training loss. Our image synthesis loss function is based on four images $I_t^l$, $I_t^r$, $I_{t+1}^l$ and $I_{t+1}^r$ and can be split into three parts
\begin{equation}
    \mathcal{L}^{\mathrm{tot}}  = \mathcal{L}^{lr} + \mathcal{L}^{\mathrm{roi}} + \mathcal{L}^{\mathrm{t}}
\end{equation}
Where $\mathcal{L}^{lr}$ is the loss for left-right consistency, $\mathcal{L}^{\mathrm{roi}}$ is the RoI based loss function and $\mathcal{L}^{\mathrm{t}}$ is the full image based loss function on flow and depth over time. To state how the three parts are defined we introduce the notation $ s \arrow t$ to indicate the warping from source $s$ to target $t$.
\begin{align}
    \mathcal{L}^{lr}  & = \mathcal{L}_\mathrm{photo}(I_t^l, \hat{I}_t^{r\arrow l}) + \mathcal{L}_\mathrm{photo}(I_{t+1}^l, \hat{I}_{t+1}^{r\arrow l}) \\
    \mathcal{L}^{\mathrm{roi}} & = \!\! \sum_j \mathcal{L}_\mathrm{photo}(I_{t}^{l,j} \!\!, \hat{I}_{t+1 \arrow t}^{l,j}) \!+ \!\mathcal{L}_\mathrm{geo}(D_{t}^{l,j}\!\!, \hat{D}_{t+1 \arrow t}^{l,j}, F_t^{lj}) \nonumber \\
\mathcal{L}^{\mathrm{t}} &= \mathcal{L}_\mathrm{photo}(I_{t}^l, \hat{I}_{t+1 \arrow t}^l) +  \mathcal{L}_\mathrm{geo}(D_t^l, \hat{D}_{t+1 \arrow t}^l, F_t^l) \nonumber 
\end{align}

Beside the loss detailed above, we use additional priors such as smoothness for depth and flow while respecting discontinuties at boundaries~\cite{godard2017unsupervised}. Optionally, we use a classical stereo system such as ELAS~\cite{Geiger2010ACCV} to compute an incomplete disparity map and use it as weak supervision in form of an $L_{\mathrm{1}}$ loss.

\begin{figure}[t]
    \centering
    \includegraphics[width=0.99\linewidth]{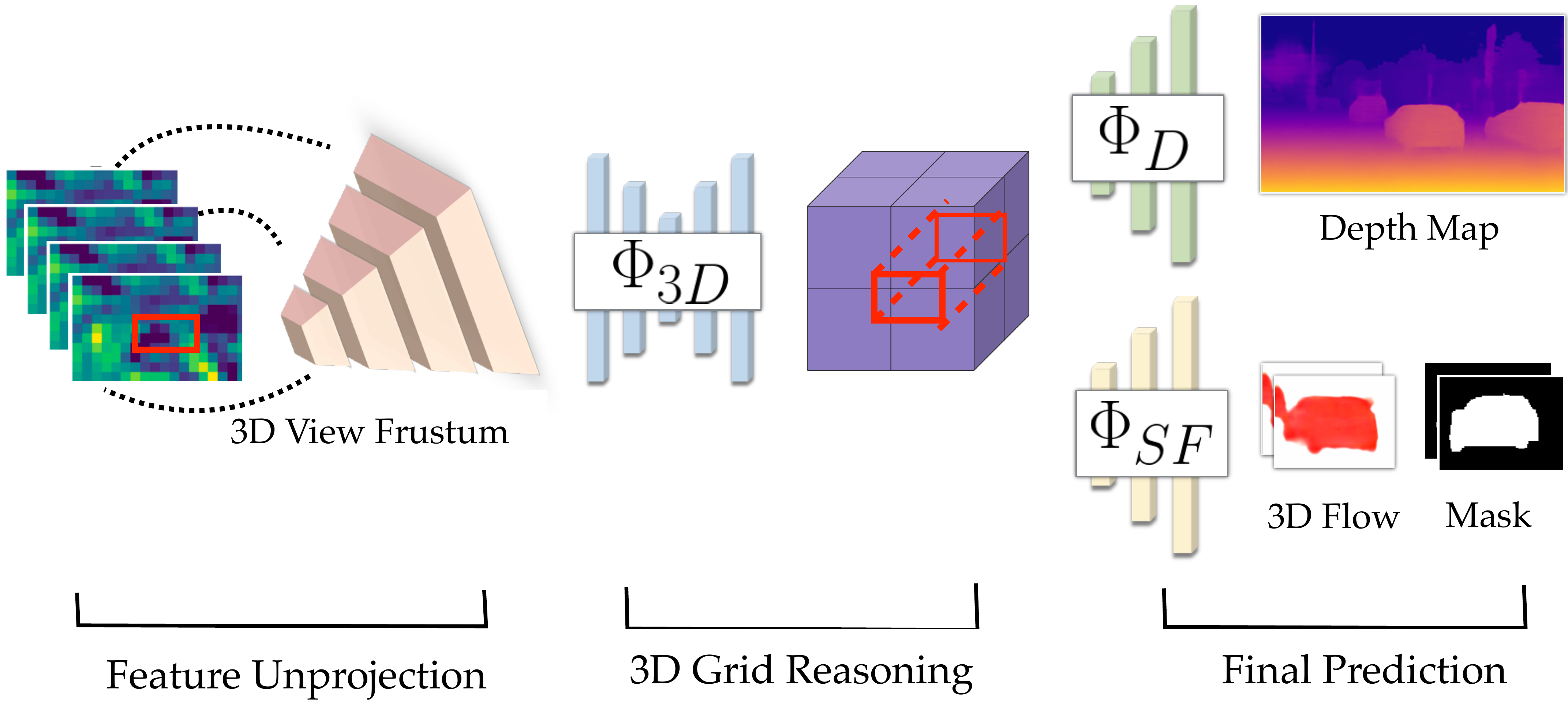} \\
    \caption{Network architecture. Our system predicts depth and instance-level 3D scene flow in a single forward pass. With extracted image features, we unproject features into a discretized view frustum grid, and then use a 3D CNN $\Phi_{3D}$ and finally perform prediction using depth $\Phi_D$ and scene flow $\Phi_{SF}$ decoders.}
    \figlabel{network}
\end{figure}


\section{Network Architecture}

\seclabel{network}
\figref{network} illustrates the our network for scene flow, mask and depth prediction. We first talk about the 3D grid representation used to integrate the information from all images and then describe each component of the network.

\subsection{3D Grid Representation}
\seclabel{3dgrid}

In order to enable the network to reason about the scene geometry in 3D, we unproject the 2D features to construct a 3D grid~\cite{kar2017learning}. A common way is using a discretization that splits a 3D cuboid volume of interest into equally sized 
\begin{wrapfigure}{r}{0.3\columnwidth}
 \vspace{-0.15in}
 \includegraphics[width=1\linewidth]{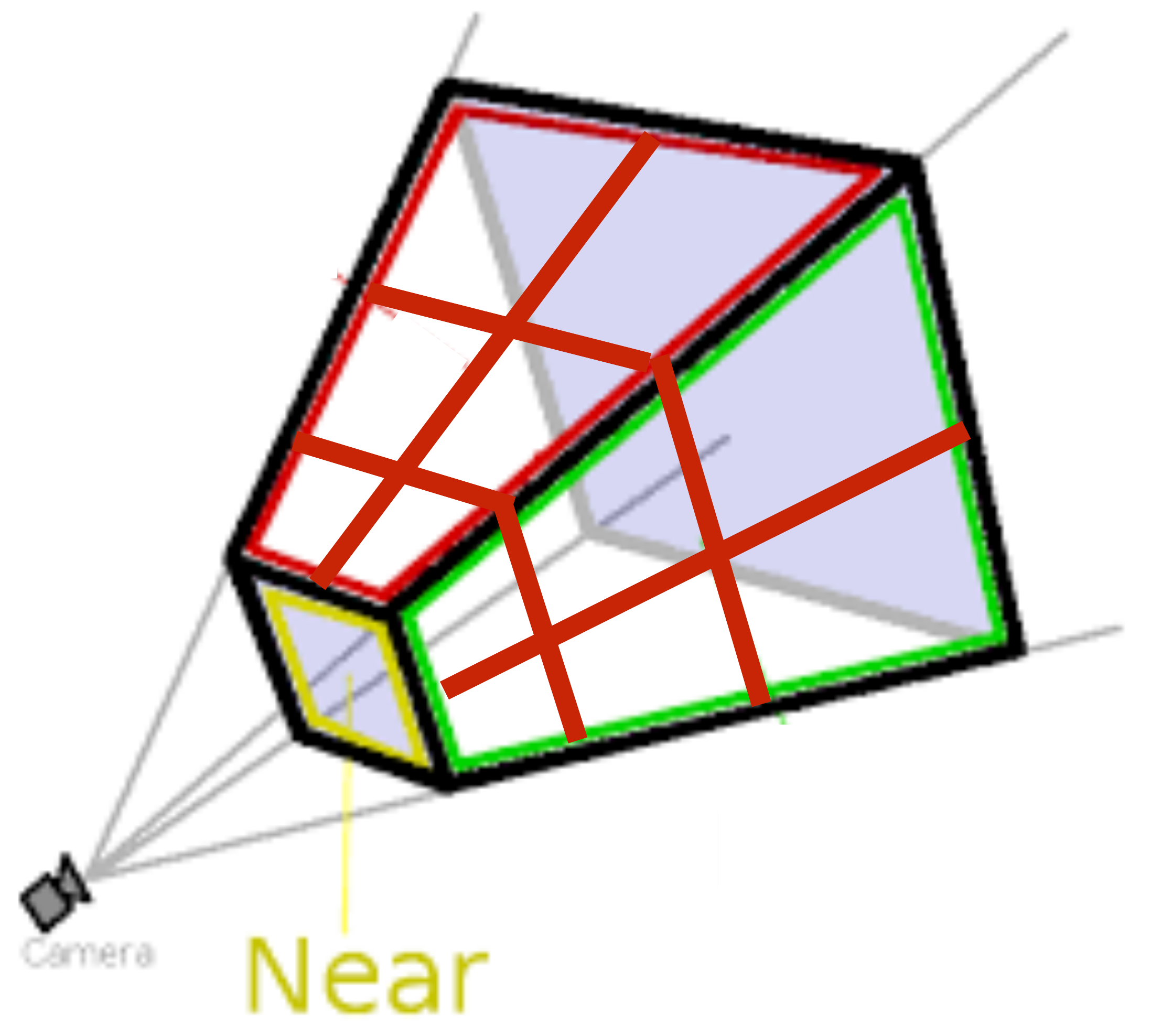}
 \vspace{-0.3in}
 \label{fig:sceneflow}
\end{wrapfigure}
voxels. This representation is used for 3D object shape reconstruction~\cite{drcTulsiani17, kar2017learning}. However, it is not suitable for the case of outdoor scenes with a large depth range, where we want to be more certain about foreground objects' geometry and motion, and allow increasing uncertainty of the depth estimates with increasing depth in the 3D world. This lends to using the well known frustum shaped grid called matching cost volume or plane sweep volume in classical (multi-view) stereo. In learning based stereo it has recently been used in~\cite{yao2018mvsnet}. The grid is discretized in image space plus an additional inverse depth ("nearness") coordinate, as shown in above image.

\subsection{Network Components}
\vspace{-3pt}
\paragraph{Image Encoder.} In a first stage the images are processed using a 2D CNN  $\Phi_I$, which outputs for each image a 2D feature map with $c$ feature channels. The weights for this CNN are shared for all input frames - typically stereo frames at two time instants $\{I_t^l, I_t^r\}$ and $\{I_{t+1}^l, I_{t+1}^r\}$.

\vspace{-10pt}
\paragraph{Unprojection.}
Using the 3D grid defined in \secref{3dgrid}, we lift the 2D information into the 3D space. We use the two left camera images as references images $\{I_t^l, I_{t+1}^l\}$ and generate these 3D grids in both their camera coordinates. Each grid is populated with image features from all 4 images by projecting the grid cell centers into the respective images using the corresponding projection matrices~\cite{kar2017learning}. We use the left images as reference frames as we predict disparity maps for both and scene flow from $I_t^l$ to $I_{t+1}^l$.

\vspace{-12pt}
\paragraph{Grid Pooling.}
The grids from the previous stage contain image features from all 4 frames. In order to combine the information from multiple frames we use two strategies. We use element-wise max pooling for features from left and right pairs and concatenate the features for different time instants in each grid cell. The motivation behind the above is that for stereo frames, there is no object motion and hence the feature should align well after unprojection. Thus a simple strategy of max pooling works well. Whereas for frames at different time instants, we expect motion in the scene and thus there would be misalignment in the image features where objects move. The output from this stage are two grids $G_t^l$ and $G_{t+1}^l$.

\vspace{-12pt}
\paragraph{3D Grid Reasoning.}
The next module $\Phi_{3D}$ processes the above two grids independently and generates output grids of the same resolution $\tilde{G}_t^l$ and $\tilde{G}_{t+1}^l$. This module is implemented as a 3D encoder-decoder CNN module with skip connections following the U-Net architecture \cite{ronneberger2015unet}.

\vspace{-12pt}
\paragraph{Output Modules.}
The final output is based on two CNN modules - one producing full frame depth for each reference image and one producing scene flow for each RoI in frame $I_t$. For each image $I_i^l$, with $i \in \{t, t+1 \}$  we first collapse $\tilde{G}_i^l$ (a 4D tensor) into a 3D tensor $C_i^l$ by concatenating features in the depth dimension. As the grid is aligned with the reference image's camera, this corresponds to accumulating the features from various disparity planes at every pixel into a single feature. This tensor is further processed using $\phi_D$ to produce the full frame disparity map. The 3D flow prediction follows an RCNN~\cite{girshick14CVPR} based architecture where given RoIs, we crop out corresponding regions $C_t^l$ using an RoI align layer~\cite{he2017mask} and pass them onto $\phi_{SF}$ which predict the scene flow and instance mask for each RoI. We also use skip connections from the image encoder in $\phi_D$ and $\phi_{SF}$ to produce sharper predictions. The full frame scene flow map is created from the RoIs by pasting back as described in \secref{roiassembly}. The final outputs from our system are disparity maps $D_t^l$ and $D_{t+1}^l$ and a forward scene flow map $F_t^l$.

\renewcommand{\arraystretch}{1.2}
\setlength{\tabcolsep}{8pt}
\begin{table*}[htb!]
\centering
\vspace{0pt}
\centering
\footnotesize
\resizebox{0.98\linewidth}{!}{
\begin{tabular*}{1.0\linewidth}{l|c|c|c|c|c|c|c|c}
\toprule
\textbf{Method}& \textbf{AMAD}$^{\downarrow}$ & \textbf{AMAE}$^{\downarrow}$ & \textbf{AE$\leq$15}$^{\circ\uparrow}$ & \textbf{AE$\leq$30}$^{\circ\uparrow}$ & \textbf{SMAD}$^{\downarrow}$ & \textbf{SMAE}$^{\downarrow}$ & \textbf{SE$\leq$0.15}$^{\uparrow}$ & \textbf{SE$\leq$0.3}$^{\uparrow}$ \\
\midrule
GeoNet~\cite{yin2018geonet} + Godard~\cite{godard2017unsupervised}  & 6.98$^\circ$ & 28.82$^\circ$ & 62.93 & 77.16 &0.256 & 0.503 & 0.351 & 0.554\\
UnflowC~\cite{meister2017unflow} + Godard~\cite{godard2017unsupervised} & 5.96$\degree$ &26.94$\degree$& 64.87&77.58 &0.240 &0.471&36.21 &58.62\\
Ours (no RoI consistency loss) & 6.03$^\circ$ & 29.34$^\circ$ & 67.59 & 75.94 & 0.207 & 0.358 & 37.46 & 58.93 \\
Our 3D scene flow & \textbf{5.19$^\circ$} & \textbf{22.92$^\circ$} & \textbf{74.78} & \textbf{78.87} & \textbf{0.193} &\textbf{0.334} & \textbf{40.95} & \textbf{62.72} \\
\bottomrule
\end{tabular*}
}
\vspace{5pt}
\caption{Comparison of instance-level object motion in terms of motion direction($A$) and speed ($S$). MAE denotes the mean average error, MAD denotes the median absolute deviation. The lower the better. We also report the percentage of the angle/speed error below different thresholds, where AE denotes the absolute angular error, SE denotes the absolute speed error. The higher the better.}
\label{tab::3dflow}
\end{table*}

\begin{figure}
\begin{center}
   \includegraphics[width=1.0\linewidth]{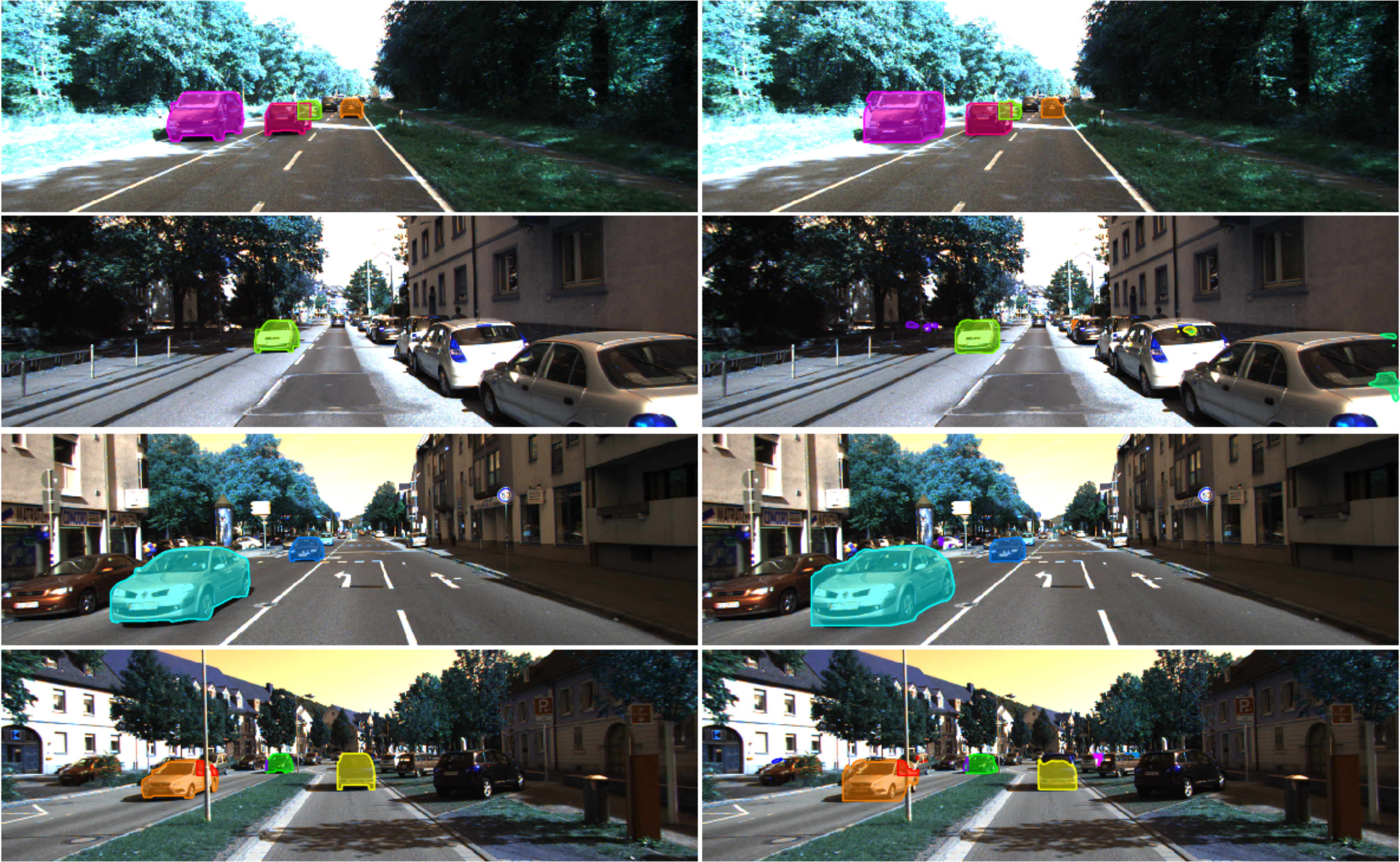}
\end{center}
\vspace{-3ex}
\hspace{28pt} {\scriptsize (a) Ground-truth} \hspace{70pt}  {\scriptsize (b) Prediction} \\
\vspace{-2.3ex}
   \caption{Qualitative results on our instance-level moving object mask prediction. Instances are color-encoded.}
   \vspace{-2ex}
\figlabel{exp_mask}
\end{figure}

\begin{figure*}
\begin{center}
   \includegraphics[width=1.\linewidth]{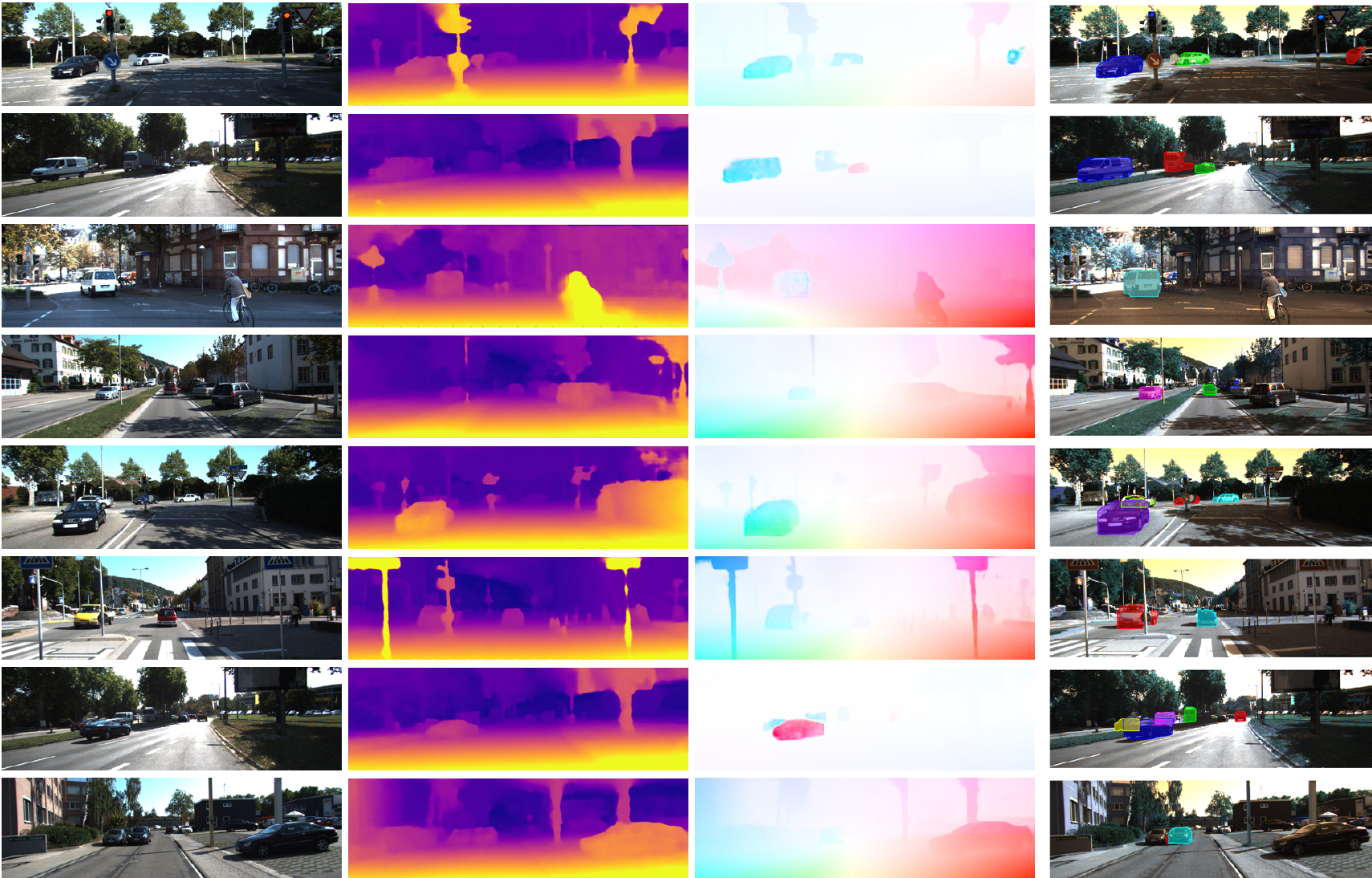}
\end{center}
\vspace{-1ex}
   \caption{Qualitative results of our method. From left to right, we show the reference image, our depth, optical flow and instance-level moving object mask prediction.}
\figlabel{fig::exp_qua}
\end{figure*}

\section{Experiments}
\seclabel{experiments}
We evaluate our instance-level 3d object motion and mask prediction on the KITTI 2015 sceneflow dataset~\cite{Menze2015CVPR}. This is the only available dataset that contains real images together with ground-truth scene flow annotations. Following existing work~\cite{meister2017unflow,yin2018geonet,zou2018df,godard2017unsupervised}, we adopt the official 200 training images as test set. The official testing set is adopted for the final finetuning process. This is possible as we do not require the ground truth for training. All the related images in the 28 scenes covered by test data are excluded for training.  \figref{exp_mask} and \figref{fig::exp_qua} show some qualitative results. 

%
\vspace{-10pt}
\paragraph{Training details} Our system is implemented using TensorFlow~\cite{abadi2016tensorflow}. All models are optimized end-to-end using Adam~\cite{kingma2014adam} with a learning rate of $1\times10^{-4}$, decay rate of 0.5 and decay steps of 100000. During training, we randomly crop the input images in the horizontal direction to obtain patches with the size of $384\times 640$ as input to the network. We set the output size of each RoI as $128\times128$, we set the number of channels in the 3D grid to 64. The batch size is set as 1 to deal with flexible RoI number for training patch. For the image encoder, we finetune the first $4$ convolutional layers from Inception Resnet V2~\cite{szegedy2017inception} pretrained on ImageNet. The rest of network is trained from scratch. We first train the depth prediction for 80K iterations on the KITTI raw dataset and then jointly train the depth and scene flow prediction for another 100k iterations. We finetune the model on the official testing set for another 120k iterations and use official 200 training images for comparison with other methods. The whole training process takes about 30 hours using a single NVIDIA Titan-X GPU. 

\subsection{Moving Object Speed and Direction Evaluation}
Our method predicts 3D sceneflow for each independently moving object.
For each test image pair, ground-truth annotation of the disparity image at time $t$, the disparity image at time $t+1$ warped into the first image's coordinate frame and the 2D optical flow from time $t$ to time $t+1$ are provided. Using these GT annotations together with the estimated camera egomotion obtained from Libviso2~\cite{Geiger2011IV}, we compute the 3D scene flow in the format of $(x,y,z)$ for each image.
To provide an instance-level analysis, we use the bbox detections~\cite{liu2016ssd}, and find the dominant 3d flow over each object region.
As a result, we represent the motion direction and speed for each instance using a single 3d flow vector in the ground truth and all algorithms. We evaluate with the following metrics: the mean average error of the euclidean length of the 3d flow (speed), the mean average error of the angle of the 3d flow (motion direction) from the moving objects pixels. To be more robust with the outliers, we also report the percentage of the mean average error below different thresholds.
In order to compare with other self-supervised flow and depth learning methods we need to reconstruct scene flow from depth and optical flow prediction. Geonet provides depthmaps with unknown scale factor and unflow does not estimate depth, we therefore use the depth results from Godard~\etal~\cite{godard2017unsupervised}. As shown in Table~\ref{tab::3dflow}, the average instance-level motion direction error of our method is less than $23^\circ$, about $15\%$ smaller than the result obtained from the best self-supervised optical flow combined with the best self-supervised depth algorithm. In our prediction, about $75\%$ of moving instances have an angular error below 15$^\circ$.

\renewcommand{\arraystretch}{1.2}
\setlength{\tabcolsep}{6pt}
\begin{table}[t]
\centering
\small
\setlength{\tabcolsep}{5.0pt}
\resizebox{0.45\textwidth}{!}{
\begin{tabular*}{1.0\linewidth}{l|c|c}
\toprule
\textbf{Method} & \textbf{Image IoU} & \textbf{Instance IoU} \\
\midrule
Zhou et al.~\cite{zhou2017unsupervised}  & 
0.380 & -\\
Bounding box detections~\cite{liu2016ssd}& 0.365 & 0.655\\
Our mask prediction & \textbf{0.624} & \textbf{0.842}\\
\bottomrule
\end{tabular*}
}
\vspace{10pt}
\caption{Moving object mask evaluation. We report IoU number in both the full image and the moving instance bounding box. }
\label{tab::mask}
\vspace{-3ex}
\end{table}

\renewcommand{\arraystretch}{1.2}
\setlength{\tabcolsep}{6pt}
\begin{table*}[!htpb]
\centering
\footnotesize
\resizebox{0.98\linewidth}{!}{
\begin{tabular*}{1.0\linewidth}{l|ccc|ccc|ccc|ccc}
\toprule
\textbf{Method} & \multicolumn{3}{c|}{\textbf{D1}}                          & \multicolumn{3}{c|}{\textbf{D2}}                          & \multicolumn{3}{c}{\textbf{FL}}  & \multicolumn{3}{c}{\textbf{ALL}}                         \\
    & bg & fg & bg+fg & bg & fg  & bg+fg  & bg  & fg & bg+fg & bg  & fg & \textbf{bg+fg}  \\ 
    \midrule
EPC~\cite{yang2018every}        & 23.62          & 27.38          & 26.81          & 18.75          & 70.89          & 60.97          & 25.34          & 28.00          & 25.74          \\ 
EPC++~\cite{luo2018every} (mono) & 30.67 & 34.38  & 32.73 & 18.36  & 84.64 & 65.63 & 17.57 & 27.30  & 19.78 & $>$30.67 & $>$84.64 & $>$65.63         \\
EPC++~\cite{luo2018every} (stereo) & 22.76 & 26.63 & 23.84 & 16.37 & 70.39 & 60.32 & 17.58  & \textbf{26.89} & \textbf{19.64} & $>$22.76 & $>$70.39 & $>$60.32\\
Godard~\etal~\cite{godard2017unsupervised} & 9.43 &18.74 & 10.86& -&-&-&-&-&-&-&-&-\\
GeoNet~\cite{yin2018geonet} & -&-&-&-&-&-& 43.54 & 48.24 &44.26&-&-&-\\
Godard~\cite{godard2017unsupervised} + GeoNet flow & 9.43 &18.74 & 10.86 & 9.10& 25.95& 25.42 &43.54 & 48.24 &44.26 &48.22 &55.75 &49.38 \\
Ours & \textbf{6.27} & \textbf{15.95} & \textbf{7.76} & \textbf{8.46} & \textbf{23.60} & \textbf{10.92} & \textbf{14.36} & 51.25 & 20.16 & \textbf{16.58} & \textbf{53.20} & \textbf{22.64} \\
\bottomrule
\end{tabular*}}
\vspace{3pt}
\caption{Results on KITTI 2015 scene flow training split. All number shows the percentage of correctly predicted pixels. D1 denotes the disparity image at time $t$, D2 denotes the disparity image at time $t+1$ warped into the first frame, FL denotes the 2D optical flow between the two time instances, fg denotes the foreground, and bg denotes the background.}
\label{tab:sceneflow}
\end{table*}

\subsection{Moving Object Instance Mask Evaluation}
Our method can produce instance-level moving object segmentation from object bounding boxes and stereo videos. This is achieved without any instance mask ground truth supervision. We evaluate our predictions on the KITTI sceneflow 2015 training split. The dataset provides an ``Object map'' which contains the foreground moving cars in each image. We use this motion mask as ground truth in our segmentation evaluation. \figref{exp_mask} shows some qualitative result of our moving object mask prediction. As shown in Table~\ref{tab::mask}, we evaluate our mask prediction using the Intersection Over Union (IoU) metric. Specifically, We compute the mean image-level IoU which considers both moving object and static background and the mean instance-level IoU for only moving objects. Our method achieves highest IoU for mask prediction. As a baseline comparison, we use mask generated from SSD~\cite{liu2016ssd} 2D bounding box detections. Those masks contain both moving and static cars, thus it can only achieve an mean IoU of 0.34 for the full image mask. Even with the GT object movement information, it does not have tight object boundary and thus can only achieve a mean IoU of 0.655. This illustrates how our method effectively learns to determine which object is moving and identify an accurate instance segmentation for moving cars. We improve the result on both image-level and instance-level IoU. We also compare with Zhou~\etal~\cite{zhou2017unsupervised} 
which generates the foreground mask for all moving objects and occlusion region in the image. Their methods do not provide instance-level information, hence we cannot obtain the instance-level IoU numbers.

\renewcommand{\arraystretch}{1.2}
\setlength{\tabcolsep}{6pt}
\begin{table}[t]
  \centering
  \small
\resizebox{0.45\textwidth}{!}{
\begin{tabular*}{1.0\linewidth}{l|c|c|c|c}
\toprule
\textbf{Method} & \textbf{Binocular} & \textbf{Abs Rel} & \textbf{Sq Rel} & \textbf{RMSE} \\
  \midrule 
  Godard et al.~\cite{godard2017unsupervised} & no  & 0.124 & 1.388 & 6.125 \\
  \midrule
  LIBELAS~\cite{Geiger2010ACCV} & yes  &  0.186 &2.192 & 6.307 \\
  Godard et al.~\cite{godard2017unsupervised} & yes  & 0.068 & 0.835 & 4.392 \\
  Ours & yes  & \textbf{0.065} & \textbf{0.699} & \textbf{3.896} \\
  \bottomrule
  \end{tabular*}
  }
  \vspace{5pt}
    \caption{Results on the KITTI 2015 stereo training set of 200 disparity images. All learning-based methods are trained on KITTI raw dataset excluding the testing image sequences. The top half shows method which uses monocular image as input, the bottom half shows methods which use binocular images as input.}
    \label{tab:kitti_stereo}
    \vspace{-2pt}
\end{table}

\renewcommand{\arraystretch}{1.2}
\setlength{\tabcolsep}{6pt}
\begin{table}[t]
\centering
\small

\begin{tabular*}{1.0\linewidth}{l|c|c|c}
\toprule
\textbf{Method}  & \textbf{Dataset} & \textbf{Non-occluded} & \textbf{All Region} \\
\midrule
EpicFlow~\cite{revaud2015epicflow} & - & 4.45 & 9.57 \\
FlowNetS~\cite{fischer2015flownet} & C+ST & 8.12 & 14.19 \\
FlowNet2~\cite{ilg2017flownet} & C+T & 4.93 & 10.06 \\
\midrule
GeoNet~\cite{yin2018geonet} & K & 8.05 & 10.81  \\
DF-Net~\cite{zou2018df} & K+SY & - & 8.98  \\
UnFlowC~\cite{meister2017unflow} & K+SY & - & 8.80  \\
Ranjan~\etal~\cite{ranjan2018adversarial} & K & - & 7.76 \\
Ours & K & 4.97 & 5.39 \\
Ours (refined)& K & \textbf{4.19} & \textbf{5.13} \\
\bottomrule
\end{tabular*}
\vspace{5pt}
\caption{Results on KITTI 2015 flow training set over non-occluded regions and overall regions. We use the average end-point error (EPE) metric to do the comparison. The classical method EpicFlow takes 16s per frame at runtime; The FlowNetS and FlowNet2 are learned with GT flow supervision. SY denotes SYNTHIA dataset~\cite{Ros_2016_CVPR}, ST denotes Sintel dataset, C denotes FlyingChairs dataset, T denotes FlyingThings3D dataset. Numbers from other methods are directly taken from the paper.}
\label{tab::flow}
\vspace{-2ex}
\end{table}

\subsection{Optical Flow Evaluation} \label{sec::exp_flow}

An additional evaluation we can conduct is to project our 3D flow predictions back to 2D to obtain nthe optical flow. As shown in Table~\ref{tab::flow}, our method achieves the lowest EPE in both non-occluded regions and overall regions and compared to other self-supervised methods. As a baseline comparison, we train a model without RoI consistency loss, which shows a decrease in performance. Optionally, we can add an optical flow refinement sub-network, to further improve our optical flow result. The subnetowrk is a unet which takes the warped image and the optical flow computed by projecting our output into image space, together with original image frames as input. This enable the network to further improve the optical flow prediction in a similar way as the architecture proposed in \cite{spynet2017}.

\subsection{Depth Evaluation}

To evaluate our depth prediction we use the KITTI 2015 stereo training set of 200 disparity images as test data and compare to other self-supervised learning and classical algorithms in Table.~\ref{tab:kitti_stereo}. We compare to algorithms that take binocular stereo as input at test time. Our method can achieve a higher accuracy as we input two consecutive binocular frames and our network also effectively manages to match over time.

\subsection{Scene Flow Evaluation}
We compare other unsupervised method in the sceneflow subset by directly using their released results or running their released code. For this benchmark, a pixel is considered to be correctly estimated if the disparity or flow end-point error is $\leq3$ pixels or $\leq5\%$. For scene flow this criterion needs to be fulfilled for two disparity maps and the flow map. As shown in Table~\ref{tab:sceneflow}, our method has an overall better accuracy than earlier self-supervised methods. 
Compared to classical approaches which optimize at test time our accuracy is still lower. However, test time optimization is in general prohibitively slow for real-time systems.


\section{Conclusion}
We have presented an end to end learned system to predict scene depth and object scene flow. Our network is trained using raw stereo sequences with off-the-shelf object detectors using image consistency as the key learning objective. Our formulation is general and can be applied in any setting where a dynamic scene is imaged by multiple cameras - e.g. a multi-view capture system~\cite{joo_iccv_2015}. In future work, we would like to extend our system to integrate longer range temporal information. We would also want it to have an emergent notion of objects to remove the dependence on pretrained object detectors. We also intend to explore general scenarios such as casual videos captures using dual camera consumer devices and leverage large scale training for a truly general purpose depth and scene flow prediction system.

{\small
\bibliographystyle{ieee}
\bibliography{egbib}

\begin{thebibliography}{10}\itemsep=-1pt

\bibitem{abadi2016tensorflow}
M.~Abadi, P.~Barham, J.~Chen, Z.~Chen, A.~Davis, J.~Dean, M.~Devin,
  S.~Ghemawat, G.~Irving, M.~Isard, et~al.
\newblock Tensorflow: A system for large-scale machine learning.
\newblock In {\em OSDI}, 2016.

\bibitem{bregler2000recovering}
C.~Bregler, A.~Hertzmann, and H.~Biermann.
\newblock Recovering non-rigid 3d shape from image streams.
\newblock In {\em CVPR}, 2000.

\bibitem{costeira1998multibody}
J.~P. Costeira and T.~Kanade.
\newblock A multibody factorization method for independently moving objects.
\newblock {\em IJCV}, 1998.

\bibitem{eigen2014depth}
D.~Eigen, C.~Puhrsch, and R.~Fergus.
\newblock Depth map prediction from a single image using a multi-scale deep
  network.
\newblock In {\em NIPS}, 2014.

\bibitem{fischer2015flownet}
P.~Fischer, A.~Dosovitskiy, E.~Ilg, P.~H{\"a}usser, C.~Hazirbas, V.~Golkov,
  P.~v.d. Smagt, D.~Cremers, and T.~Brox".
\newblock Flownet: Learning optical flow with convolutional networks.
\newblock In {\em ICCV}, 2015.

\bibitem{fragkiadaki2014grouping}
K.~Fragkiadaki, M.~Salas, P.~Arbelaez, and J.~Malik.
\newblock Grouping-based low-rank trajectory completion and 3d reconstruction.
\newblock In {\em NIPS}, 2014.

\bibitem{furukawa2010accurate}
Y.~Furukawa and J.~Ponce.
\newblock Accurate, dense, and robust multiview stereopsis.
\newblock {\em TPAMI}, 2010.

\bibitem{garg2016unsupervised}
R.~Garg, V.~K. BG, G.~Carneiro, and I.~Reid.
\newblock Unsupervised cnn for single view depth estimation: Geometry to the
  rescue.
\newblock In {\em ECCV}, 2016.

\bibitem{Geiger2010ACCV}
A.~Geiger, M.~Roser, and R.~Urtasun.
\newblock Efficient large-scale stereo matching.
\newblock In {\em ACCV}, 2010.

\bibitem{Geiger2011IV}
A.~Geiger, J.~Ziegler, and C.~Stiller.
\newblock Stereoscan: Dense 3d reconstruction in real-time.
\newblock In {\em Intelligent Vehicles Symposium (IV)}, 2011.

\bibitem{girshick14CVPR}
R.~Girshick, J.~Donahue, T.~Darrell, and J.~Malik.
\newblock Rich feature hierarchies for accurate object detection and semantic
  segmentation.
\newblock In {\em CVPR}, 2014.

\bibitem{godard2017unsupervised}
C.~Godard, O.~Mac~Aodha, and G.~Brostow.
\newblock Unsupervised monocular depth estimation with left-right consistency.
\newblock In {\em CVPR}, 2017.

\bibitem{joo_iccv_2015}
L.~T. L. G. B. N. I. M. T. K. S.~N. Hanbyul~Joo, Hao~Liu and Y.~Sheikh.
\newblock Panoptic studio: A massively multiview system for social motion
  capture.
\newblock In {\em ICCV}, 2015.

\bibitem{hartley2003multiple}
R.~Hartley and A.~Zisserman.
\newblock {\em Multiple view geometry in computer vision}.
\newblock Cambridge university press, 2003.

\bibitem{he2017mask}
K.~He, G.~Gkioxari, P.~Doll{\'a}r, and R.~Girshick.
\newblock Mask r-cnn.
\newblock In {\em ICCV}, 2017.

\bibitem{huguet2007variational}
F.~Huguet and F.~Devernay.
\newblock A variational method for scene flow estimation from stereo sequences.
\newblock In {\em ICCV}, 2007.

\bibitem{ilg2017flownet}
E.~Ilg, N.~Mayer, T.~Saikia, M.~Keuper, A.~Dosovitskiy, and T.~Brox.
\newblock Flownet 2.0: Evolution of optical flow estimation with deep networks.
\newblock In {\em CVPR}, 2017.

\bibitem{jaimez2015primal}
M.~Jaimez, M.~Souiai, J.~Gonzalez-Jimenez, and D.~Cremers.
\newblock A primal-dual framework for real-time dense rgb-d scene flow.
\newblock In {\em ICRA}, 2015.

\bibitem{ji2017surfacenet}
M.~Ji, J.~Gall, H.~Zheng, Y.~Liu, and L.~Fang.
\newblock Surfacenet: An end-to-end 3d neural network for multiview stereopsis.
\newblock In {\em ICCV}, 2017.

\bibitem{kar2017learning}
A.~Kar, C.~H{\"a}ne, and J.~Malik.
\newblock Learning a multi-view stereo machine.
\newblock In {\em NIPS}, 2017.

\bibitem{kendall2017end}
A.~Kendall, H.~Martirosyan, S.~Dasgupta, P.~Henry, R.~Kennedy, A.~Bachrach, and
  A.~Bry.
\newblock End-to-end learning of geometry and context for deep stereo
  regression.
\newblock In {\em ICCV}, 2017.

\bibitem{kingma2014adam}
D.~P. Kingma and J.~Ba.
\newblock Adam: A method for stochastic optimization.
\newblock {\em arXiv preprint arXiv:1412.6980}, 2014.

\bibitem{ladicky2014pulling}
L.~Ladicky, J.~Shi, and M.~Pollefeys.
\newblock Pulling things out of perspective.
\newblock In {\em CVPR}, 2014.

\bibitem{liu2016ssd}
W.~Liu, D.~Anguelov, D.~Erhan, C.~Szegedy, S.~Reed, C.-Y. Fu, and A.~C. Berg.
\newblock Ssd: Single shot multibox detector.
\newblock In {\em ECCV}, 2016.

\bibitem{longuet1981computer}
H.~C. Longuet-Higgins.
\newblock A computer algorithm for reconstructing a scene from two projections.
\newblock {\em Nature}, 1981.

\bibitem{luo2018every}
C.~Luo, Z.~Yang, P.~Wang, Y.~Wang, W.~Xu, R.~Nevatia, and A.~Yuille.
\newblock Every pixel counts++: Joint learning of geometry and motion with 3d
  holistic understanding.
\newblock {\em arXiv preprint arXiv:1810.06125}, 2018.

\bibitem{mayer2016large}
N.~Mayer, E.~Ilg, P.~Hausser, P.~Fischer, D.~Cremers, A.~Dosovitskiy, and
  T.~Brox.
\newblock A large dataset to train convolutional networks for disparity,
  optical flow, and scene flow estimation.
\newblock In {\em CVPR}, 2016.

\bibitem{meister2017unflow}
S.~Meister, J.~Hur, and S.~Roth.
\newblock Unflow: Unsupervised learning of optical flow with a bidirectional
  census loss.
\newblock {\em AAAI}, 2018.

\bibitem{Menze2015CVPR}
M.~Menze and A.~Geiger.
\newblock Object scene flow for autonomous vehicles.
\newblock In {\em CVPR}, 2015.

\bibitem{ozden2004reconstructing}
K.~E. Ozden, K.~Cornelis, L.~Van~Eycken, and L.~Van~Gool.
\newblock Reconstructing 3d trajectories of independently moving objects using
  generic constraints.
\newblock {\em CVIU}, 2004.

\bibitem{pollefeys2004visual}
M.~Pollefeys, L.~Van~Gool, M.~Vergauwen, F.~Verbiest, K.~Cornelis, J.~Tops, and
  R.~Koch.
\newblock Visual modeling with a hand-held camera.
\newblock {\em IJCV}, 2004.

\bibitem{spynet2017}
A.~Ranjan and M.~J. Black.
\newblock Optical flow estimation using a spatial pyramid network.
\newblock In {\em CVPR}, 2017.

\bibitem{ranjan2018adversarial}
A.~Ranjan, V.~Jampani, K.~Kim, D.~Sun, J.~Wulff, and M.~J. Black.
\newblock Adversarial collaboration: Joint unsupervised learning of depth,
  camera motion, optical flow and motion segmentation.
\newblock {\em arXiv preprint arXiv:1805.09806}, 2018.

\bibitem{revaud2015epicflow}
J.~Revaud, P.~Weinzaepfel, Z.~Harchaoui, and C.~Schmid.
\newblock Epicflow: Edge-preserving interpolation of correspondences for
  optical flow.
\newblock In {\em CVPR}, 2015.

\bibitem{ronneberger2015unet}
O.~Ronneberger, P.~Fischer, and T.~Brox.
\newblock U-net: Convolutional networks for biomedical image segmentation.
\newblock In {\em MICCAI}, 2015.

\bibitem{Ros_2016_CVPR}
G.~Ros, L.~Sellart, J.~Materzynska, D.~Vazquez, and A.~M. Lopez.
\newblock The synthia dataset: A large collection of synthetic images for
  semantic segmentation of urban scenes.
\newblock In {\em CVPR}, 2016.

\bibitem{szegedy2017inception}
C.~Szegedy, S.~Ioffe, V.~Vanhoucke, and A.~A. Alemi.
\newblock Inception-v4, inception-resnet and the impact of residual connections
  on learning.
\newblock In {\em AAAI}, volume~4, page~12, 2017.

\bibitem{taniai2017fast}
T.~Taniai, S.~N. Sinha, and Y.~Sato.
\newblock Fast multi-frame stereo scene flow with motion segmentation.
\newblock In {\em CVPR}, 2017.

\bibitem{tulsiani2017multi}
S.~Tulsiani, T.~Zhou, A.~A. Efros, and J.~Malik.
\newblock Multi-view supervision for single-view reconstruction via
  differentiable ray consistency.
\newblock In {\em CVPR}, 2017.

\bibitem{drcTulsiani17}
S.~Tulsiani, T.~Zhou, A.~A. Efros, and J.~Malik.
\newblock Multi-view supervision for single-view reconstruction via
  differentiable ray consistency.
\newblock In {\em CVPR}, 2017.

\bibitem{ullman1984maximizing}
S.~Ullman.
\newblock Maximizing rigidity: The incremental recovery of 3-d structure from
  rigid and nonrigid motion.
\newblock {\em Perception}, 1984.

\bibitem{vedula1999three}
S.~Vedula, S.~Baker, P.~Rander, R.~Collins, and T.~Kanade.
\newblock Three-dimensional scene flow.
\newblock In {\em ICCV}. IEEE, 1999.

\bibitem{vijayanarasimhan2017sfm}
S.~Vijayanarasimhan, S.~Ricco, C.~Schmid, R.~Sukthankar, and K.~Fragkiadaki.
\newblock Sfm-net: Learning of structure and motion from video.
\newblock Technical report, arXiv:1704.07804, 2017.

\bibitem{vogel2013piecewise}
C.~Vogel, K.~Schindler, and S.~Roth.
\newblock Piecewise rigid scene flow.
\newblock In {\em ICCV}, 2013.

\bibitem{wedel2008efficient}
A.~Wedel, C.~Rabe, T.~Vaudrey, T.~Brox, U.~Franke, and D.~Cremers.
\newblock Efficient dense scene flow from sparse or dense stereo data.
\newblock In {\em ECCV}, 2008.

\bibitem{yang2018every}
Z.~Yang, P.~Wang, Y.~Wang, W.~Xu, and R.~Nevatia.
\newblock Every pixel counts: Unsupervised geometry learning with holistic 3d
  motion understanding.
\newblock {\em arXiv preprint arXiv:1806.10556}, 2018.

\bibitem{yao2018mvsnet}
Y.~Yao, Z.~Luo, S.~Li, T.~Fang, and L.~Quan.
\newblock Mvsnet: Depth inference for unstructured multi-view stereo.
\newblock In {\em ECCV}, 2018.

\bibitem{yin2018geonet}
Z.~Yin and J.~Shi.
\newblock Geonet: Unsupervised learning of dense depth, optical flow and camera
  pose.
\newblock In {\em CVPR}, 2018.

\bibitem{yuan20063d}
C.~Yuan and G.~Medioni.
\newblock 3d reconstruction of background and objects moving on ground plane
  viewed from a moving camera.
\newblock In {\em CVPR}, 2006.

\bibitem{zhou2017unsupervised}
T.~Zhou, M.~Brown, N.~Snavely, and D.~G. Lowe.
\newblock Unsupervised learning of depth and ego-motion from video.
\newblock In {\em CVPR}, 2017.

\bibitem{zou2018df}
Y.~Zou, Z.~Luo, and J.-B. Huang.
\newblock Df-net: Unsupervised joint learning of depth and flow using
  cross-task consistency.
\newblock In {\em ECCV}, 2018.

\end{thebibliography}
}

\end{document}